\documentclass[letterpaper, 10 pt, conference]{ieeeconf}  

\IEEEoverridecommandlockouts                              

\usepackage{amsmath} 
\usepackage{amssymb}  
\usepackage{graphicx} 
\usepackage{float}
\usepackage{booktabs}
\usepackage{algorithm}
\usepackage{algorithmic}
\usepackage{xcolor}
\usepackage{makecell}
\usepackage{multirow}
\usepackage{cite}
\usepackage{mathtools}
\usepackage{url}

\title{\LARGE \bf
FastStair: Learning to Run Up Stairs with Humanoid Robots
}

\author{
Yan Liu$^{1,2}$,
Tao Yu$^{2}$, 
Haolin Song$^{4}$,
Hongbo Zhu$^{2, 5}$,
Nianzong Hu$^{2}$, 
Yuzhi Hao$^{2, 6}$, 
Xiuyong Yao$^{2}$,\\
Xizhe Zang$^{1 \text{\textdagger}}$, 
Hua Chen$^{2,3 \text{\textdagger}}$,
Jie Zhao$^{1}$
\thanks{* Work done at LimX Dynamics. Project lead: Tao Yu}
\thanks{$\text{\textdagger}$ Corresponding author}
\thanks{$^1$School of Mechanics Engineering, Harbin Institute of Technology (HIT), Harbin Heilongjiang 150001, China. Email: {\tt\footnotesize liuyan98@stu.hit.edu.cn, zangxizhe@hit.edu.cn, jzhao@hit.edu.cn}}
\thanks{$^2$ LimX Dynamics, Shenzhen, China.}
\thanks{$^3$ Zhejiang University-University of Illinois Urbana-Champaign Institute (ZJUI), Haining, China. Email: {\tt\footnotesize huachen@intl.zju.edu.cn}}
\thanks{$^4$ Department of Electronic Engineering and Information Science
(EEIS), University of Science and Technology of China, Hefei 230027,
China. Email: {\tt\footnotesize hlsong@mail.ustc.edu.cn}}
\thanks{$^5$ Hong Kong University of Science and Technology, Hong Kong SAR, China. Email: Email: {\tt\footnotesize hzhubi@connect.ust.hk}}
\thanks{$^6$ Department of Mechanical Engineering, National University of Singapore, Singapore 117575. Email: {\tt\footnotesize  yuzhihao@u.nus.edu}}
}

\begin{document}

\maketitle
\thispagestyle{empty}
\pagestyle{empty}

\begin{abstract}

Running up stairs is effortless for humans but remains extremely challenging for humanoid robots due to the simultaneous requirements of high agility and strict stability. Model-free reinforcement learning (RL) can generate dynamic locomotion, yet implicit stability rewards and heavy reliance on task-specific reward shaping tend to result in unsafe behaviors, especially on stairs; conversely, model-based foothold planners encode contact feasibility and stability structure, but enforcing their hard constraints often induces conservative motion that limits speed. We present FastStair, a planner-guided, multi-stage learning framework that reconciles these complementary strengths to achieve fast and stable stair ascent. FastStair integrates a parallel model-based foothold planner into the RL training loop to bias exploration toward dynamically feasible contacts and to pretrain a safety-focused base policy. 
To mitigate planner-induced conservatism and the discrepancy between low- and high-speed action distributions, the base policy was fine-tuned into speed-specialized experts and then integrated via Low-Rank Adaptation (LoRA) to enable smooth operation across the full commanded-speed range.
We deploy the resulting controller on the Oli humanoid robot, achieving stable stair ascent at commanded speeds up to 1.65 m/s and traversing a 33-step spiral staircase (17 cm rise per step) in 12 s, demonstrating robust high-speed performance on long staircases. Notably, the proposed approach served as the champion solution in the Canton Tower Robot Run Up Competition. Project Page: \url{https://npcliu.github.io/FastStair}.

\end{abstract}

\section{INTRODUCTION}
Humanoid robots have attracted significant interest because of their superior terrain adaptability, yet they remain far from matching human agile and stable locomotion. Staircases, a ubiquitous feature of human environments, present a particular challenge. While humans navigate them intuitively, robots face an inherent trade-off between agility and stability that remains unresolved. This trade-off has plagued state-of-the-art locomotion control methods, hindering the deployment of humanoid robots in real-world stair-climbing tasks.

Reinforcement Learning (RL) has emerged as a powerful paradigm for generating dynamic humanoid locomotion \cite{li2025reinforcement, liao2025beyondmimic, peng2025gait}. Although RL can achieve impressive speeds on flat terrain \cite{li2025reinforcement}, its application to challenging terrains like staircases often exposes a fragility. This vulnerability stems from the reliance on implicit stability rewards, where dynamic stability is not explicitly guaranteed but is instead encouraged through a ensemble of heuristic reward terms, such as alive, posture, and gravity-projection. On stairs, these implicit objectives frequently conflict with task-level rewards such as high-velocity tracking. This inherent contradiction forces the learning process into a trade-off where the pursuit of agility often leads the robot to sacrifice the very rewards that maintain stability. Consequently, most current stair-climbing control schemes, including both "blind" \cite{gu2024humanoid, wang2024toward} and perception-augmented \cite{long2025learning, sun2025learning, gadde2025no, ren2025vb} approaches, tend to prioritize safety by heavily weighting stability terms over speed. This prioritization effectively forfeits the potential for high-speed locomotion and results in the conservative behaviors typical of existing RL-based stair climbers.

\begin{figure}[tb]
    \centering    \includegraphics[width=0.95\linewidth]{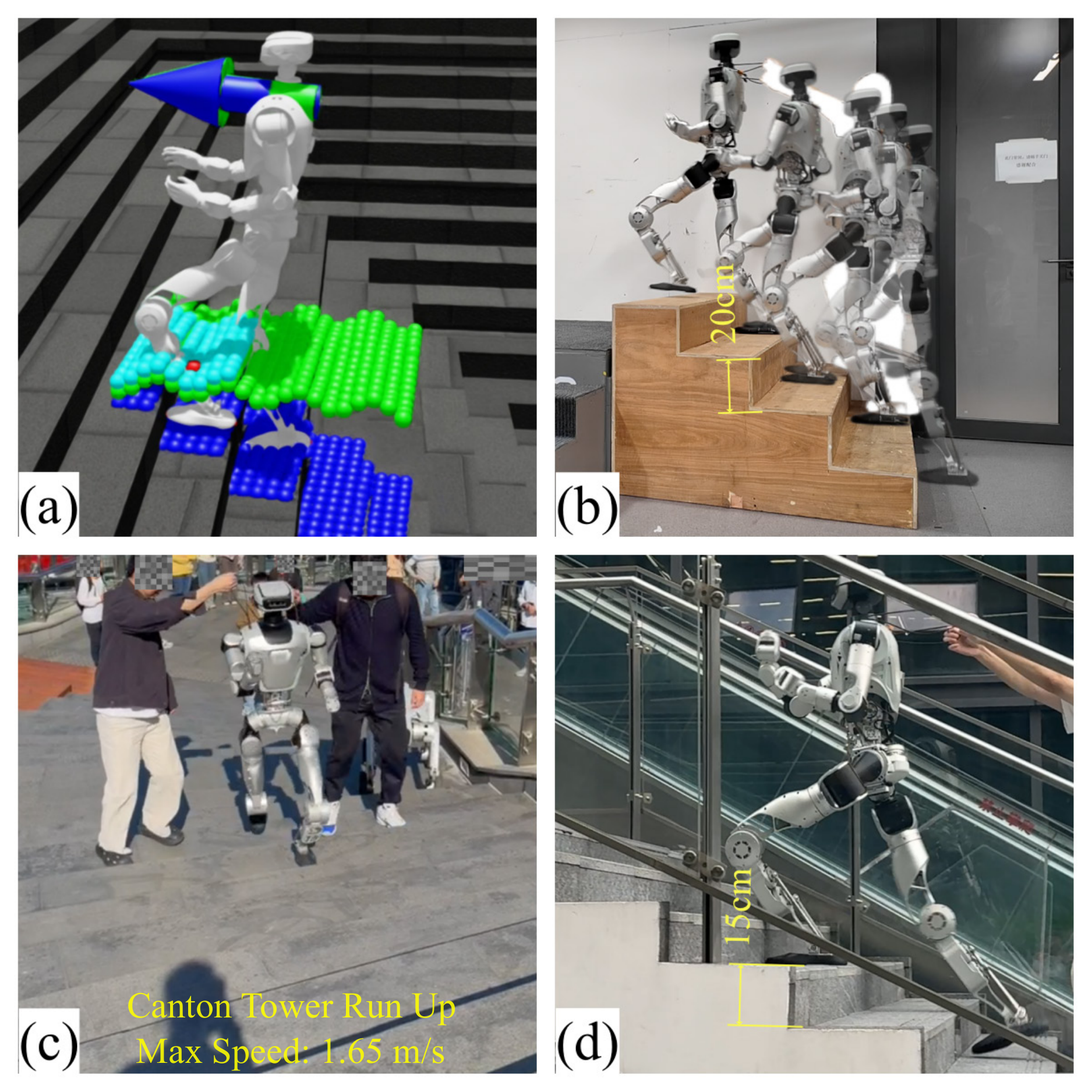}
    \setlength{\belowcaptionskip}{-15pt}  
    \caption{Full-scale humanoid robot Oli performing agile stair-climbing locomotion: (a) parallel foothold planning in IsaacLab, (b) climbing a staircase with 20\,cm step height, (c) Canton Tower Run-up, and (d) high-speed ascent.}
    \label{fig:robot_real_1}
\end{figure}

In contrast, model-based planner provides explicit stability guarantees by computing feasible footholds under model and environment constraints \cite{acosta2023bipedal, penco2025anticipatory, shi2025bipedal, acosta2025perceptive}. By enforcing hard constraints on capture regions and footstep placement, the planner ensure a high degree of safety. However, this rigorous adherence to feasibility constraints often induces conservative motion, which narrows the exploration space and limits the robot’s dynamic agility. 

A natural strategy to bridge this gap is to utilize model-based planners to guide RL exploration toward explicitly safe and feasible contacts. However, such guidance often transfers planner's inherent conservatism to the learned policy, resulting in behaviors that are safe but lack the requisite agility for high-speed tasks. This persistent tension between planner-guided safety and learned agility motivates the development of a framework that can reconcile these complementary strengths.

In this work, we propose FastStair, a planner-guided, multi-stage learning framework that reconciles model-based stability with learned agility to achieve high-speed stair-climbing locomotion.

First, regarding the foothold planner, compared with mainstream ALIP-based MPC methods \cite{acosta2023bipedal, acosta2025perceptive}, planners based on the Divergent Component of Motion (DCM) admit analytical expressions \cite{khadiv2020walking} and permit relatively simple constraint formulations. These properties make DCM-based planners readily convertible into discrete-search formulations that are well suited for GPU-parallel computation and therefore appropriate for guiding RL training.

Second, to resolve the inefficiency of implicit safety exploration, we integrate a parallel DCM-based foothold planner \cite{khadiv2020walking} into the RL loop. By reformulating the optimization as a parallel discrete search, we avoid reliance on parallel real-time optimization solvers and guide exploration toward dynamically feasible safety regions. This "optimization-as-search" approach provides the real-time guidance necessary to pretrain a safety-focused base policy with negligible computational overhead.

Finally, to mitigate planner-induced conservatism, we adopt a multi-stage training framework. Owing to the discrepancy in action distributions between low- and high-velocity motion, a single policy struggles to track the full commanded-speed range efficiently. Accordingly, in the second stage, the base policy was fine-tuned into two speed-specialized experts via velocity-related reward reweighting. However, direct switching between these experts causes instability (e.g., chattering). To address this, in the third stage, the parameters of both experts were integrated into a single network with low-rank adaptation (LoRA) layers on their branches. The integrated model was then fine-tuned over the full commanded speed range, yielding a unified policy that transitions smoothly across full speed range and enables stable, high-speed stair climbing.

Our main contributions can be summarized as follows:
\begin{itemize}
\item To enhance stability on challenging staircases, we introduce a DCM-based foothold planner as an explicit stability reward, which guides the reinforcement learning exploration toward dynamically feasible footholds on challenging stairs.
\item To mitigate the prohibitive training overhead of incorporating real-time optimization into parallel RL, we reformulate the foothold optimization as a parallelized discrete search, accelerating the training speed by approximately 25 times.
\item To mitigate planner-induced conservatism and improve velocity-tracking performance, we propose a multi-stage framework that fine-tunes high-speed and low-speed experts from a safety-focused base policy, achieving high-speed locomotion without sacrificing stability or low-speed tracking accuracy. 

\item To address instabilities arising from expert switching, the experts were merged into a single network and fine-tuned using LoRA, which enabled smooth transitions between experts and produced a unified policy that operates robustly across the full commanded-speed range.
\end{itemize}

\begin{figure*}[htb]
    \centering
    \includegraphics[width=0.85\textwidth]{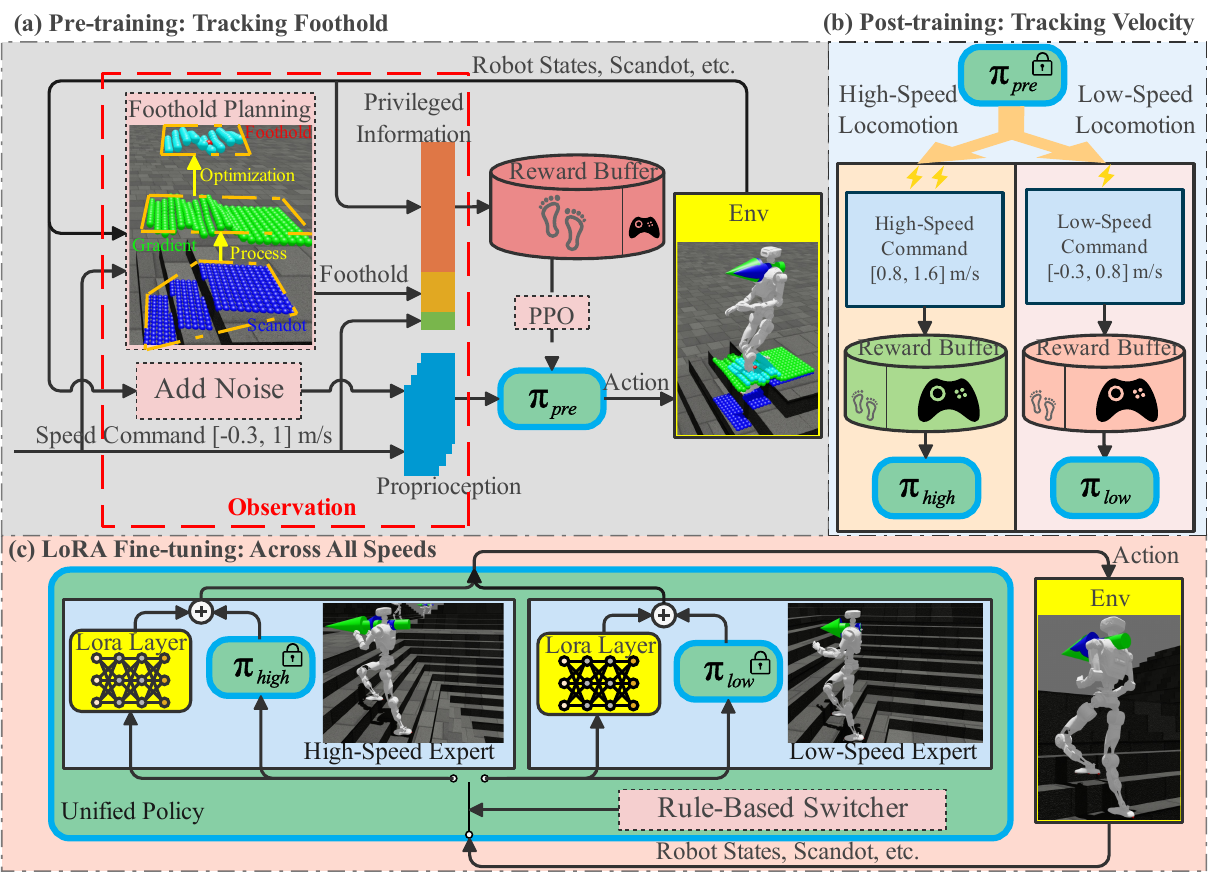}
    \setlength{\belowcaptionskip}{-10pt}  
    \caption{Overview of the FastStair framework. During the  pre-training stage, a parallel DCM-based foothold optimizer generates dynamically feasible contacts to guide policy learning via a foothold-tracking reward. In the post-training stage, the pre-trained base model is fine-tuned into high- and low-speed expert policies by expanding the commanded velocity range and adjusting reward weights to mitigate planner-induced conservatism; this decomposition is motivated by the distinct action distributions at high versus low speeds. However, direct switching between experts can cause control discontinuities, so in the LoRA fine-tuning stage we merge their parameters into a single network and fine-tune it with LoRA. The resulting unified policy ensures robust tracking across the full velocity range for deployment on the physical robot.}
    \label{fig:pipeline}
\end{figure*}

\section{Related Work}
\subsection{Model-based Locomotion on Challenging Terrains}
Classical model-based methods, such as those based on the DCM \cite{khadiv2020walking} and Model Predictive Control (MPC), provide a rigorous mathematical foundation for foothold planning \cite{acosta2023bipedal, shi2025bipedal, shamsah2023integrated}. By encoding contact feasibility and balance criteria as explicit kinemodynamic constraints, these methods excel at maintaining stability on discrete terrains like stairs or stepping stones. However, ensuring real-time computational feasibility often necessitates simplifying assumptions, such as the linear inverted pendulum model, alongside the use of conservative safety margins. These constraints inevitably restrict the reachable solution space, inducing the conservative locomotion that hinder high-speed agility. 

\subsection{Learning-based Humanoid Locomotion on Stairs}
Recent advancements in deep reinforcement learning (RL) have significantly enhanced the terrain adaptability of humanoid robots. Residual-based methods \cite{duan2024learning, gadde2025no, ren2025vb} typically employ a two-stage pipeline where a vision-based network provides corrective residuals to a frozen blind policy. While effective, these decoupled architectures often require complex coordination and may limit the robot's peak agility due to the restricted search space of the residual. Conversely, end-to-end frameworks \cite{long2025learning, sun2025learning} unify perception and control, but they often struggle with the severe exploration bottlenecks of pure RL, especially when traversing discrete terrains like stairs without explicit motion priors.

To further enhance locomotion performance, researchers have explored specialized training strategies. For instance, \cite{wang2025beamdojo} uses a curriculum based on scan points, while \cite{wang2025more} leverages human motion retargeting to induce natural behaviors. However, these methods often fail to ensure strict dynamical feasibility at high speeds. Our work diverges by using a multi-stage training framework—transitioning from safety-critical pre-training to specialized experts. By integrating these experts into a single network via LoRA \cite{hu2022lora}, we achieve high-speed stair climbing that surpasses the conservative limits of single-policy RL.

\section{Method}
We propose a multi-stage training framework, FastStair, designed to enable high-speed stair climbing while ensuring robust traversal success rates. An overview is depicted in Fig.~\ref{fig:pipeline}.
In the pre-training stage, we incorporate parallel DCM-based foothold planning to guide the policy via a large foothold-tracking reward.
While this prioritization ensures safety, it inevitably compromises velocity tracking accuracy. To address this, the post-training stage rebalances reward weights to prioritize velocity tracking and trains separate high-speed and low-speed expert policies, initialized from the pre-trained model.
Finally, in the fine-tuning stage, these experts are integrated into a network and fused via LoRA. This yields a single unified policy capable of reliable traversal across the full commanded velocity range. The details of this three-stage procedure are described below.

\subsection{Parallel Foothold Planning}\label{sec:footholdplan}

Traversing complex terrain requires selecting safe, flat footholds. However, acquiring this capability is often challenging for pure RL. To bridge this gap, we incorporate a model-based planner to guide foothold selection.
While conventional planning methods \cite{dosunmu2023stair, acosta2023bipedal} demonstrate robust performance, they typically rely on serial CPU-based optimization solvers, rendering them incompatible with massive parallel RL frameworks. Even recent advancements in parallel optimization \cite{jeon2024cusadi} often incur non-trivial computational overhead when scaled.
To address this, we formulate a DCM-based foothold optimization problem and reformulate it as a parallel discrete search task. By solving this task via vectorized tensor operations, we leverage the GPU's massive parallelism, ensuring the planner incurs negligible computational overhead during training.

\subsubsection{DCM for Stair Climbing}
\begin{figure}[tb]
    \centering    
    \includegraphics[width=0.8\columnwidth]{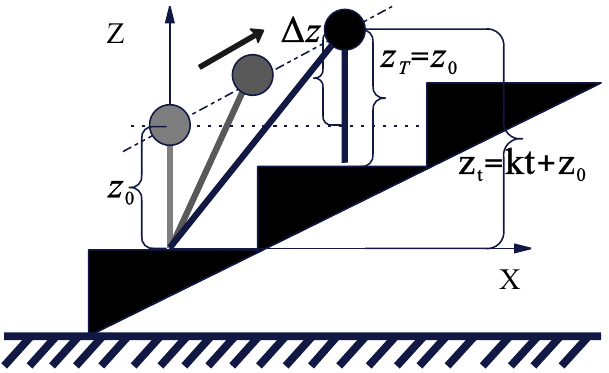}
    \setlength{\belowcaptionskip}{-15pt} 
    \caption{Inverted-pendulum stair-climbing process.}
    \label{fig:stairLIP}
\end{figure}

Following \cite{caron2019capturability}, we model the system dynamics using the Variable-Height Inverted Pendulum (VHIP), governed by: $\ddot{x} = \omega^2 x,$ 
where $x \in \mathbb{R}$ denotes position in the forward direction.  To accommodate stair climbing, we define the Center of Mass (CoM) height trajectory $z(t)$ during the swing phase (duration $T$) as a linear function $z(t) = k t + z_0$, where $z_0$ represents the nominal standing height and $k$ denotes the vertical slope. Consequently, the natural frequency is given by $\omega = \sqrt{g/z(t)}$, where $g$ is the gravitational acceleration. Fig.~\ref{fig:stairLIP} illustrates this VHIP formulation applied to stair negotiation.
The DCM is defined as $\xi = x + \frac{1}{\omega}\dot{x}$. Differentiating with respect to time yields: $\dot{\xi} = \left(1 - \frac{\dot{\omega}}{\omega^2}\right)\dot{x} + \frac{1}{\omega}\ddot{x}.$ 
Noting that $\dot{\omega} = -\frac{1}{2}\omega\frac{k}{z}$, 
then:
\begin{equation}
    \label{eq:dotdcm}
    \begin{aligned}
        \dot{\xi} &= \left(1 + \frac{k}{2\sqrt{gz}}\right)\dot{x} + \frac{1}{\omega}\ddot{x} \\
        &= a\dot{x} + \omega x \\
        &= a\omega\xi + (1-a)\omega x,
    \end{aligned}
\end{equation}
where $a = 1 + \frac{k}{2\sqrt{g z}}$. Considering a maximum stair incline of $\pm 45^\circ$ (implying $k\in[-1,1]$) and a nominal standing height $z_0 \approx 1$\,m, the coefficient $a$ remains within the range $[0.84, 1.16]$. Given that $a$ is close to unity, we approximate $a \approx 1$ to simplify the optimization. This yields the dynamics $\dot{\xi} \approx \omega\xi$ with a time-varying $\omega$. Consequently, the DCM evolution for stair climbing at time $t \in [0, T]$ is given by:$\xi(t) = \xi_0 e^{\int_{0}^{t}\omega(\tau)d\tau}.$

\subsubsection{DCM-based Foothold Optimization}
\begin{figure}[tb]
    \centering
    \includegraphics[width=0.6\linewidth]{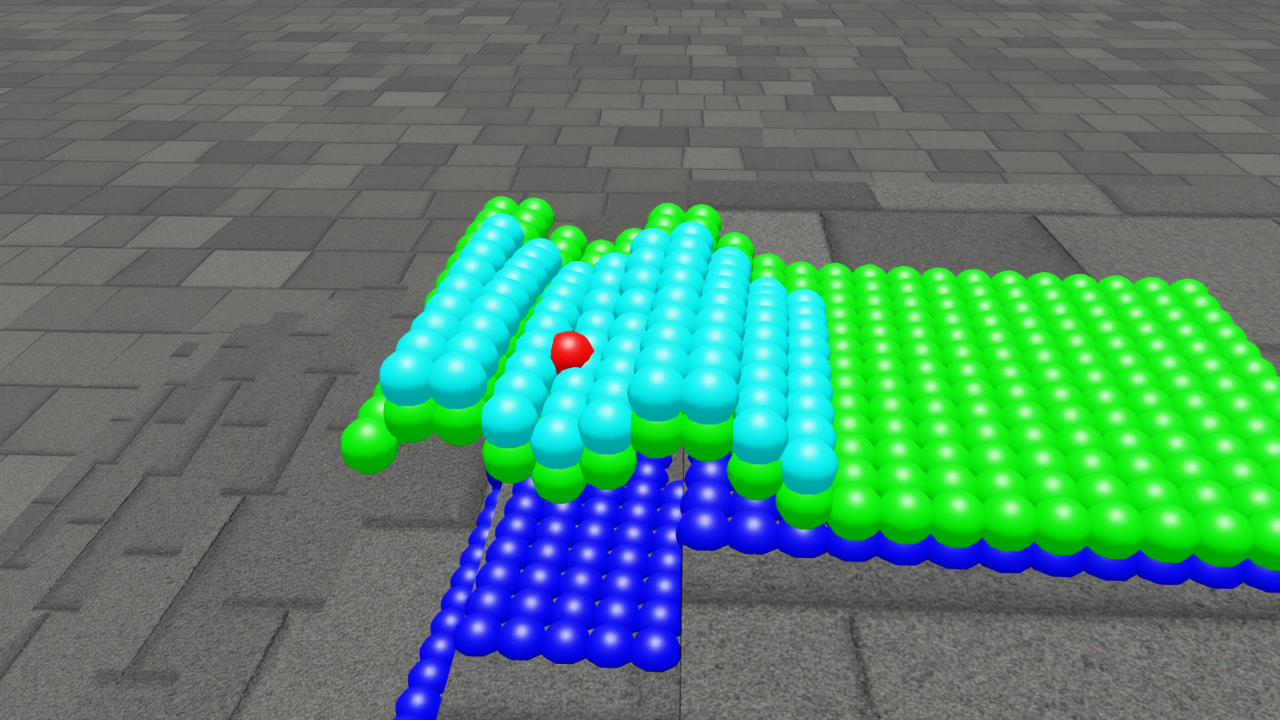}
    \caption{Terrain-perception information. Blue markers show terrain scandots, green the averaged gradient map, light blue the local averaged gradient map, and red the optimal footholds.}
    \label{fig:scandot1}
\end{figure}

Leveraging the structural similarity between our derived dynamics and the standard planar DCM model, we adapt the optimization formulation from \cite{khadiv2020walking}:
\begin{align}
    \min_{u_T, b} \quad & \alpha_1 \left\| u_T - \begin{bmatrix} L_{\text{nom}} \\ W_{\text{nom}} \end{bmatrix} \right\|^2 + \alpha_2 \| b - b_{\text{nom}} \|^2 + \alpha_3 S(u_T) \label{eq:objective} \\
    \text{s.t.} \quad & u_T + b = \xi_t e^{\sigma(T) - \sigma(t)}, \label{eq:constraint3} \\
    & u_T \in \mathcal{P}, \label{eq:constraint4}
\end{align}
The integrated natural frequency is explicitly given by:$\sigma(t) = \int_{0}^{t} \omega(\tau)d\tau = \frac{2\sqrt{g}\left(\sqrt{k t + z_0} - \sqrt{z_0}\right)}{k}.$
The decision variable $u_T \in \mathbb{R}^2$ represents the planar projection of the candidate foothold, while $b$ denotes the DCM offset \cite{khadiv2020walking}. The nominal step targets are defined as $L_{\text{nom}} = v_x^d T$ and $W_{\text{nom}} = v_y^d T + (-1)^i l$, where $(v_x^d, v_y^d)$ is the desired commanded velocity, $i$ is the leg index (left/right), and $l$ denotes the nominal lateral inter-foot distance. Accordingly, the nominal DCM offset is formulated as $b_{\text{nom}}=[\frac{L}{e^{\sigma(T)} - 1}, (-1)^i \frac{l_p}{1 + e^{\sigma(T)}} - \frac{W}{1 - e^{\sigma(T)}}]$. The feasible foothold set $\mathcal{P}$ is derived from a local elevation map (spanning $1.8\,\text{m} \times 1.2\,\text{m}$) centered on the robot, as visualized by the blue points in Fig.~\ref{fig:scandot1}. The term $S(u_T)$ quantifies the terrain steepness cost (indicated by light-blue points in Fig.~\ref{fig:scandot1}).

\subsubsection{Steepness Score Calculation}
The steepness score $S(u_T)$ evaluates local terrain gradients to identify feasible contact patches. Upon acquiring the elevation map, we compute the spatial gradient using a Sobel operator, implemented efficiently as a GPU-parallelized convolution. To account for the robot's planar foot morphology, we apply a subsequent local neighborhood aggregation step (also via convolution) over a kernel matching the foot dimensions. In the resulting cost map, higher values correspond to steeper terrain, indicating reduced suitability for foothold placement.

\subsubsection{Optimization-to-Search Reformulation}
Directly solving large-scale optimization problems in real-time typically incurs prohibitive computational overhead, creating a bottleneck for parallel RL. To overcome this, we reformulate the optimization into a parallel discrete search.
We observe that the feasible foothold set $\mathcal{P}$ (blue points in Fig.~\ref{fig:scandot1}) is naturally discretized by the elevation map resolution and can be encoded as a tensor on the GPU. Leveraging the equality constraint in Eq.~\eqref{eq:constraint3}, we analytically map each candidate foothold $u_T \in \mathcal{P}$ to its corresponding DCM offset $b$. Consequently, the cost function in Eq.~\eqref{eq:objective} can be evaluated in parallel across the entire batch of candidate footholds using efficient vectorized tensor operations. The optimal foothold is then identified via a simple \texttt{argmin} reduction.
To further accelerate the query, we constrain the search window to a $0.6\,\text{m} \times 0.6\,\text{m}$ region of interest centered on the nominal target $[L_{\text{nom}}, W_{\text{nom}}]^\top$, as indicated by the light-blue region in Fig.~\ref{fig:scandot1}.

While this formulation relies on simplified dynamics (approximating $a \approx 1$), it provides a computationally lightweight yet effective heuristic sufficient to guide the RL policy. Moreover, this \textit{optimization-to-search} paradigm offers a generalizable strategy for tightly coupling model-based planners with massive parallel RL environments without sacrificing training speed.

\subsection{Three-Stage Reinforcement Learning Framework}
\subsubsection{Observation Space}
The policy's state space is partitioned into proprioceptive observations (available during both training and deployment) and privileged observations (restricted to the training phase).
The proprioceptive observations primarily comprise the user command vector $\boldsymbol{c}_t=[v_x^d, v_y^d, w_{\mathrm{yaw}}^d]^\top$, base angular velocity, projected gravity vector, joint positions, joint velocities, and the previous action history. To encode timing, we include gait clock signals $\{\sin(2\pi ft), \cos(2\pi ft)\}$, where $f$ denotes the stepping frequency. Crucially, the robot also perceives the terrain via elevation map centered on the robot.

The privileged observations augment this set with ground-truth states and explicit model-based guidance. These include the base linear velocity, base height, external contact forces, joint accelerations, and joint torques. Importantly, to guide the learning process, we incorporate the optimal foothold computed by the parallel planner described in Sec.~\ref{sec:footholdplan}.

\subsubsection{Reward Function and Network Architecture}
Since our primary contribution lies in the multi-stage framework, we adopt the actor-critic network architecture and the baseline reward formulation from \cite{GaitAdaptive2025Haolin}. However, to drive stage-specific behaviors, we adjust specific reward scalars across the three training stages, as detailed in Table~\ref{tab:reward}.

Crucially, to enforce the guidance provided by the parallel planner, we introduce a specific foothold-tracking reward not present in \cite{GaitAdaptive2025Haolin}: $r_{\text{foot}} = \exp\!\left(-10 \| \mathbf{p}_f - \mathbf{p}_f^d \| \right),$ where $\mathbf{p}_f \in \mathbb{R}^3$ denotes the current swing-foot position and $\mathbf{p}_f^d \in \mathbb{R}^3$ represents the desired instantaneous position for the next time step. The desired trajectory $\mathbf{p}_f^d$ is generated via interpolation (e.g., B\'{e}zier curves) defined by three keyframes: the lift-off position, the swing apex, and the target landing position $[u_T^\top, h_z]^\top$. Here, $u_T$ is the optimal planar foothold derived in Sec.~\ref{sec:footholdplan}, and $h_z$ is the corresponding terrain height retrieved directly from the elevation map. To prevent tripping, the swing apex is configured to ensure sufficient clearance above the maximum height difference between the lift-off and landing points.

\subsubsection{Pre-training Stage}
In the Pre-training Stage, we prioritize traversal stability over precise velocity tracking. Accordingly, we assign a dominant weight to the foothold-tracking reward defined above. This constraint compels the policy to target safe contact regions—specifically the centers of stair treads—rather than risky edge placements, thereby maximizing the probability of successful traversal. Through this process, the policy effectively encodes the geometric intuition of the model-based planner. However, the emphasis on foothold accuracy leads to increased speed-tracking error.

\subsubsection{Post-training Stage}\label{sec:Posttraining}
To recover velocity-tracking precision and increase climbing speed, the reward weights were adjusted by increasing the weight of the speed-tracking term and reducing the foothold-tracking weight. 

We partition the command space into two distinct regimes: a high-velocity band ($[0.8, 1.6]\,\text{m/s}$) and a low-velocity band ($[-0.3, 0.8]\,\text{m/s}$). Initialized from the pre-trained policy, two separate expert model are trained to master each specific regime. This divide-and-conquer strategy is essential; empirical results indicate that training a single policy directly over the full union of ranges ($[-0.3, 1.6]\,\text{m/s}$) leads to mode collapse towards moderate velocities, significantly degrading tracking accuracy at the operational extremes (i.e., very slow or very fast motion).

\subsubsection{LoRA Fine-tuning Stage}
Because joint-action distributions differ between high- and low-speed locomotion, directly switching between discrete expert policies induces control discontinuities and physical instability during deployment. To resolve this, we integrate the parameters of both the high-speed and low-speed experts into a unified network (see Fig. \ref{fig:pipeline}). We then fine-tune the model using LoRA to smooth the transition boundaries and ensure robust performance across the full command spectrum.

Within the unified network, we employ a rule-based switcher driven by the user command: the high-speed expert is activated to generate joint actions when the commanded forward velocity $v_x^d > 0.8\,\text{m/s}$, while the low-speed expert governs the regime below this threshold.

\begin{table}[htb]
\centering
\caption{The reward weights for the three stages}
\label{tab:reward}
\setlength{\tabcolsep}{3pt}
\renewcommand{\arraystretch}{1.1}
\begin{tabular}{lccc}
\toprule
Reward Term & Stage 1 Weights  & Stage 2 Weights & Stage 3 Weights \\
\midrule
Lin. velocity track & 1.0 & 1.35 & 1.1 \\
\makecell[l]{Base height \\ (w.r.t.\ feet) }
& 0.4 & 0.3 & 0.4 \\
\midrule
Action smoothness & $-2.5$e-03 & $-2$e-03 & $-2$e-03  \\
Joint accel.\ L2 & $-5$e-07  & $-4$e-07 & $-4$e-07  \\
Joint torque L2 & $-4$e-07  & $-3$e-07 & $-3$e-07 \\
Torque rate & $-1.5$e-07  & $-2$e-08 & $-2$e-08 \\
Joint power & $-2.5$e-07  & $-2$e-07 & $-2$e-07 \\
\midrule
Lin.\ accel.\ L2 & $-2$e-03 & $-1.5$e-03 & $-1.5$e-03 \\
Proj.\ gravity L2 & $-0.15$ & $-0.1$ & $-0.1$ \\
\midrule
Feet stumble & $-1.5$ & $-1.5$ & $-2$ \\
Foothold & $0.6$ & $0.5$ & $0.4$ \\
\bottomrule
\end{tabular}
\vspace{-2mm}
\end{table}

\subsection{Elevation Maps Reconstruction}\label{TerrainScandotsReconstruction}
The locomotion policy relies on egocentric local elevation maps (visualized as blue markers in Fig.~\ref{fig:scandot1}) for terrain awareness. During simulation training, we extract ground-truth height samples directly from the IsaacLab interface. 

For physical deployment, we implement the perception pipeline proposed in \cite{GaitAdaptive2025Haolin} using an Intel RealSense D435i depth camera. This module processes raw depth frames in real-time to reconstruct a local terrain grid covering a $1.8\,\text{m} \times 1.2\,\text{m}$ area centered on the robot, with a spatial resolution of $5\,\text{cm}$. The resulting terrain scandots serves as the primary exteroceptive input for the onboard policy.

\section{EXPERIMENTS}
\subsection{Robot Platform}
\begin{figure}[t]
    \centering
    \includegraphics[width=0.3\columnwidth]{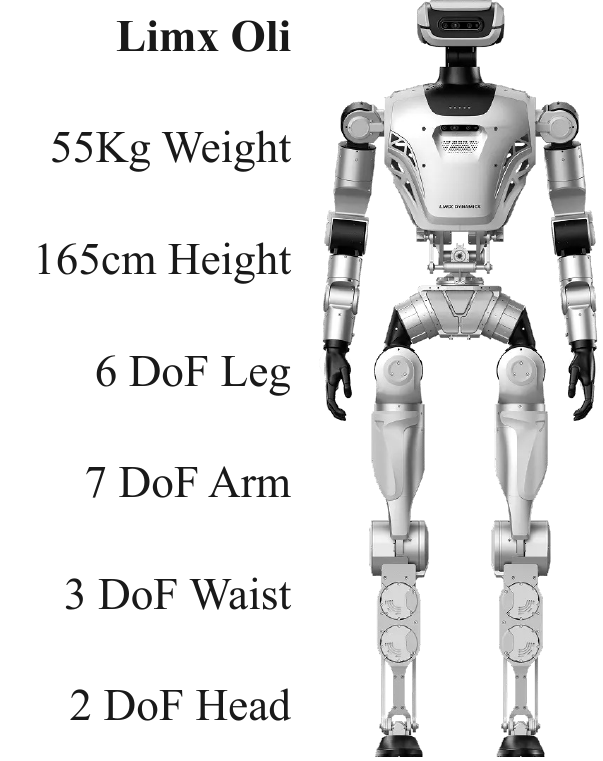}
    \setlength{\belowcaptionskip}{-15pt} 
    \caption{Robot Platform Limx Oli.}
    \label{fig:robot}
\end{figure}

We conduct all experiments using the full-scale humanoid robot LimX Oli \cite{oli} (Fig.~\ref{fig:robot}). The robot weighs 55\,kg and stands 1.65\,m tall, featuring 31 actuated degrees of freedom (DoF): 6 per leg, 7 per arm, 3 in the waist, and 2 in the head. In this study, all 31 joints are actively controlled. For onboard computing, the system integrates an NVIDIA Jetson Orin NX and a Rockchip RK3588-based single-board computer to handle real-time perception and policy inference.

\subsection{Traning and Deployment}
Following \cite{GaitAdaptive2025Haolin}, we utilize IsaacLab to construct a diverse training curriculum featuring flat ground, rough terrain, and pyramid-shaped stairs (both ascent and descent). To facilitate agile maneuvers, we constrain the gait frequency to $[1.0, 1.5]$\,Hz. We simulate 4096 environments in parallel, randomizing the robot's base pose upon each reset and resampling command velocities every 10 seconds. While the forward velocity range varies across the three training stages (as detailed in Sec. \ref{sec:Posttraining}), the lateral velocity and yaw rate are fixed at $v_y^d \in [-0.3, 0.3]$\,m/s and $w_{\text{yaw}}^d \in [-1.0, 1.0]$\,rad/s, respectively.

To bridge the sim-to-real gap, we apply domain randomization. Randomized parameters include the mass and inertia of links (base, waist, legs), contact friction, restitution coefficients, and joint PD gains. Additionally, we inject random external force and impulse perturbations. For the final LoRA fine-tuning stage we configure the LoRA hyperparameters as $\alpha=16$ and $\gamma=8$.

For real-world deployment, we employ a distributed computing architecture. The Intel RealSense D435i captures depth frames, which are processed by the perception pipeline on the NVIDIA Jetson Orin NX to generate local elevation maps. These maps are transmitted to the Rockchip RK3588-based computer, which executes the final unified policy. This decoupling of perception and control workloads ensures efficient onboard processing and maintains the strict control loop timing required for stable agile locomotion.

\subsection{Foothold Tracking and Computational Efficiency}
\subsubsection{Foothold Tracking}
\begin{figure}[tb]
    \centering
    \includegraphics[width=0.8\linewidth]{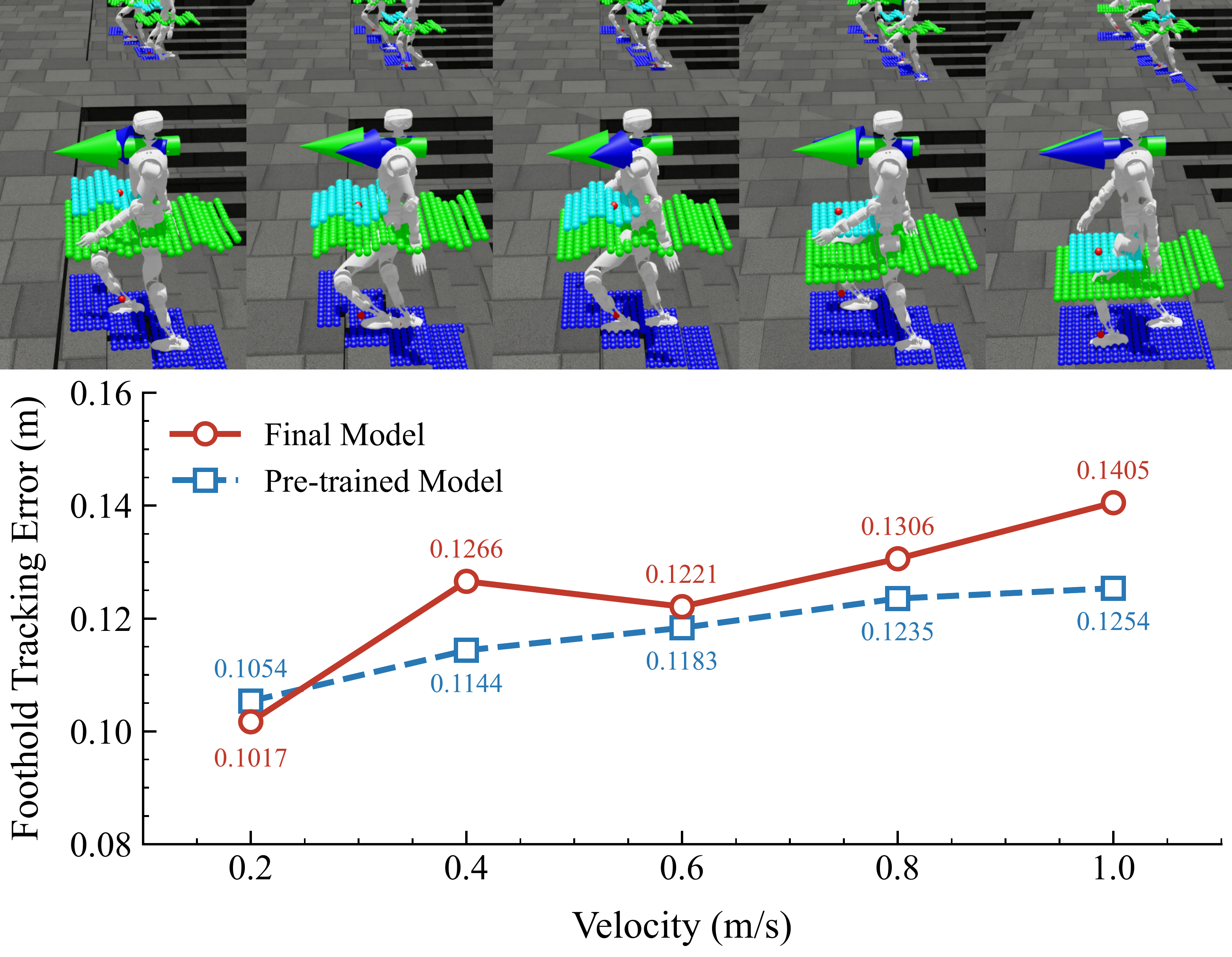}
    \setlength{\belowcaptionskip}{-15pt} 
    \caption{Foothold screenshots and Foothold-tracking error. For each commanded speed, the error was computed by averaging the absolute foothold-position errors recorded over a time window.}
    \label{fig:footstep_error}
\end{figure}

In our framework, planned footholds serve as supervisory signals within the reward function, guiding the policy toward safe contact regions. Fig.~\ref{fig:footstep_error} illustrates the optimal footholds (red point in the figure) generated via Eq.~\ref{eq:objective} during ascent; notably, the planner concentrates targets near the centers of the stair treads. By minimizing the tracking error against these targets, the robot implicitly learns to identify and exploit safe, planar contact surfaces.

To evaluate this guidance, both the pre-trained and the final unified policies were deployed in IsaacLab and data were collected at commanded forward velocities of \(\{0.4,\,0.6,\,0.8,\,1.0\}\,\text{m/s}\). Performance was quantified by the mean planar Euclidean distance between executed and planned footholds. As illustrated in Fig.~\ref{fig:footstep_error}, simulation screenshots show that the swing foot generally lands near the desired foothold; nevertheless, the foothold-tracking error of the final unified policy is larger than that of the pre-trained policy. By contrast, the velocity-tracking analysis in Fig.~\ref{fig:speed_error_mse} indicates that the final unified policy attains lower velocity error. This pattern is explained by the training design: during pre-training a larger weight is placed on foothold accuracy to prioritize safety, and only after a safe baseline has been learned is post-training used to explore more agile behaviors. The results therefore reflect the intended trade-off between safety and agility embodied by the proposed multi-stage framework.

\subsubsection{Computational Efficiency}

The complete planning pipeline—including elevation map processing and the parallel discrete search—incurs a latency of approximately 4\,ms per step for a batch of 4096 environments on an NVIDIA RTX 4090. This represents an approximate $25\times$ speedup compared to recent parallel MPC implementations using CusADi \cite{kamohara2025rl} under similar hardware conditions, highlighting the scalability of our approach for large-scale RL.

\subsection{Full-Range Velocity-Tracking Experiments}
\begin{figure}[tb]
    \centering
    \includegraphics[width=0.8\linewidth]{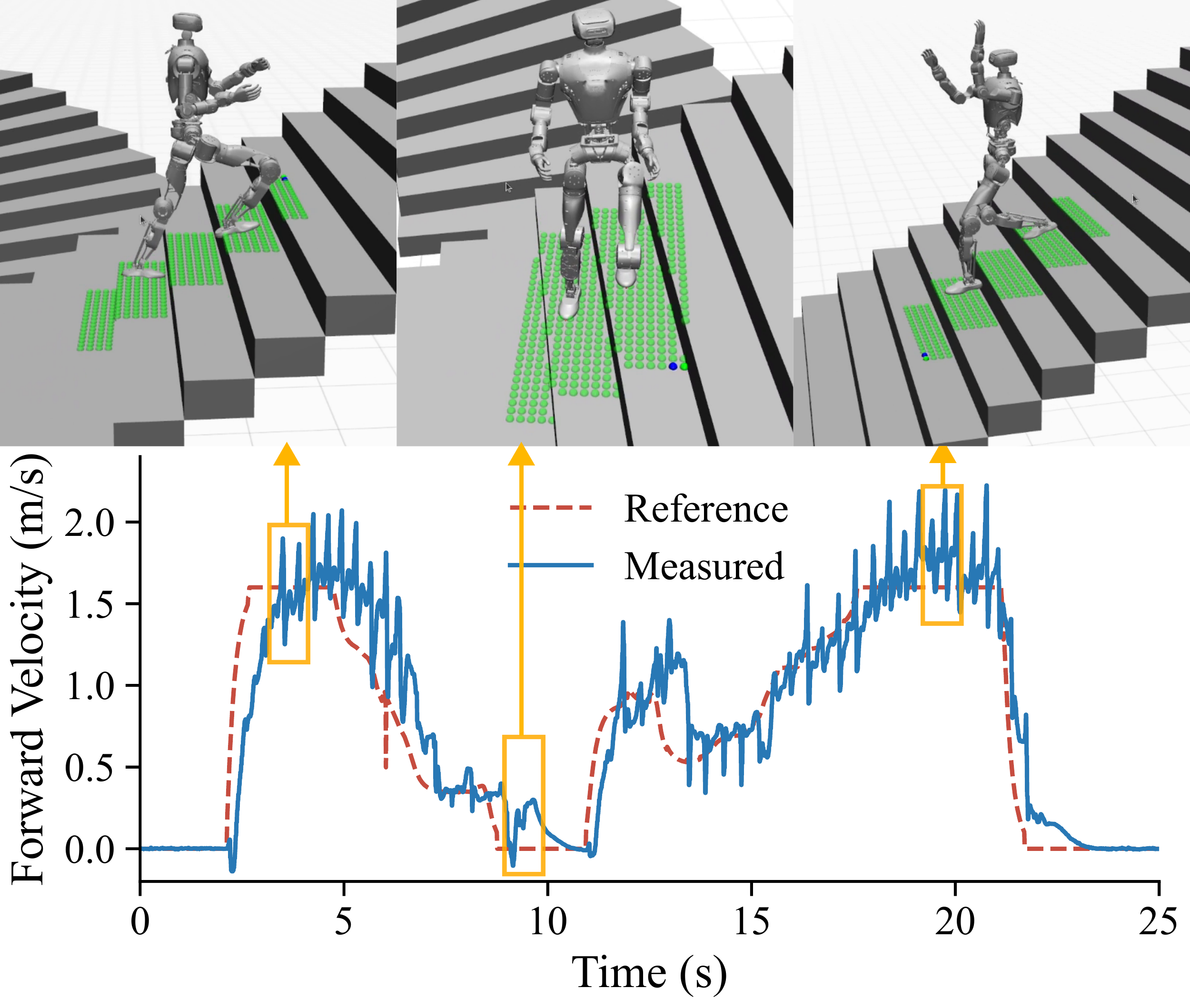}
    \setlength{\belowcaptionskip}{-15pt} 
    \caption{MuJoCo simulation snapshot and velocity-tracking performance during stair descent. The plot shows commanded and measured forward velocities over time.}
    \label{fig:Velocity_Tracking}
\end{figure}

To assess policy robustness beyond the training simulator, we conduct validation in MuJoCo on a staircase with a step height of $15\,\text{cm}$. Fig.~\ref{fig:Velocity_Tracking} presents both a simulation snapshot and a time-series plot of the forward velocity response during a stair descent maneuver. 

Experimental results demonstrate that the unified policy effectively tracks velocity commands across the entire operational range. Notably, the robot maintains stability and tracking accuracy from a stationary stance ($0\,\text{m/s}$) up to the maximum commanded velocity of $1.6\,\text{m/s}$, validating the effectiveness of the multi-stage training framework.

\subsection{Training Framework Evaluation}

We evaluated the following methods in terms of terrain traversal success ratio and velocity-tracking mean absolute error (MAE):
\begin{itemize}
  \item \textbf{Baseline \cite{GaitAdaptive2025Haolin}:} A standard end-to-end RL policy trained without model-based foothold guidance, which has twice the parameters of a velocity expert policy.
  \item \textbf{Pre-trained (Ablation):} An intermediate variant (Stage 1 only) incorporates DCM-based foothold guidance and uses twice the parameters of a velocity expert, but lacks the velocity-specific split and LoRA integration.
  \item \textbf{AMP \cite{wang2025more}:} A policy trained using Adversarial Motion Priors (AMP), where exploration is guided by retargeted human reference motions rather than terrain-aware geometric planning.
  \item \textbf{FastStair (ours):} The complete proposed framework (see Fig. \ref{fig:pipeline}). 
\end{itemize}

\subsubsection{Terrain Traversal Success Ratio}\label{sec:TraversalSuccessRatio}
\begin{figure}[tb]
    \centering
    \includegraphics[width=0.85\linewidth]{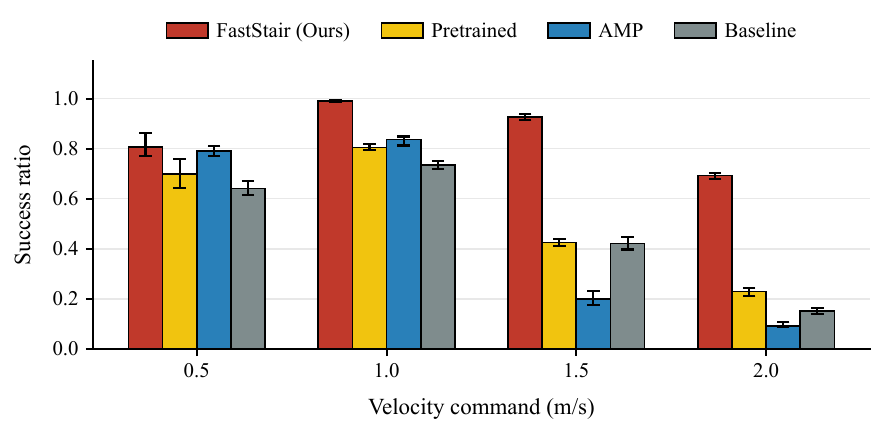}
    \setlength{\belowcaptionskip}{-15pt} 
    \caption{Success-rate comparison for traversing a staircase with 25\,cm step height at various commanded speeds.}
    \label{fig:success_ratio}
\end{figure}

The traversal success ratio was evaluated on a staircase with 25\,cm step height at commanded forward speeds of 0.5, 1.0, 1.5, and 2.0\,m/s. In IsaacSim we initialized 500 robots around the center of a pyramidal staircase with randomized headings and commanded them to traverse in all directions. The success ratio is defined as the fraction of environments in which the robot reached the far edge of the terrain within an episode without falling. Results are presented in Fig.~\ref{fig:success_ratio}.

As illustrated, FastStair achieves the highest success rate across all tested velocities. The Pre-trained policy also significantly outperforms the baselines, a result we attribute to the parallel foothold planning incorporated during pre-training. This explicitly imparts geometric guidance, enabling the robot to identify optimal contact regions on the stair treads.

In contrast, while the AMP method performs well at lower velocities ($\le 1.0\,\text{m/s}$), its performance degrades sharply at higher speeds, eventually underperforming even the unguided Baseline. In simulation, we observed that the AMP policy exhibited a tendency to stall at the base of the stairs when commanded above $1.0\,\text{m/s}$. We hypothesize this stems from two factors: the lack of high-speed climbing examples in the reference dataset (distributional shift) and a conservative reward structure that penalizes the deviations required for aggressive maneuvers.

Although all policies were trained with a maximum commanded velocity of $1.6\,\text{m/s}$, we extended the evaluation to $2.0\,\text{m/s}$ to assess robustness. Notably, FastStair maintains a success rate exceeding $70\%$ even in this out-of-distribution regime, demonstrating superior generalization of the learned agile locomotion skills.
\subsubsection{Velocity-Tracking Evaluation}
\begin{figure}[tb]
    \centering
    \includegraphics[width=0.9\linewidth]{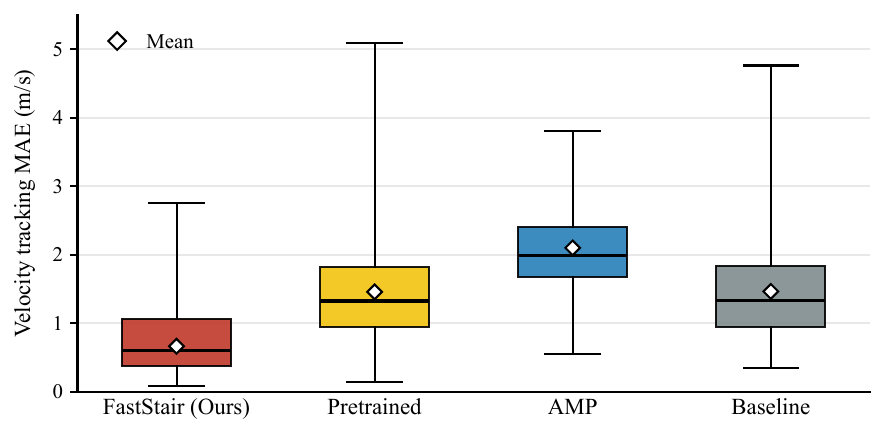}
    \setlength{\belowcaptionskip}{-15pt} 
    \caption{Mean absolute error (MAE) of velocity tracking for four methods on a staircase with 25\,cm step height.}
    \label{fig:speed_error_mse}
\end{figure}

We evaluate the velocity-tracking fidelity of the four methods on a staircase with a $25\,\text{cm}$ step height. forward velocity commands are sampled uniformly from $[-0.3, 1.6]\,\text{m/s}$ at $3\,\text{s}$ intervals. For each trial, we compute the Mean Absolute Error (MAE) between the commanded and actual velocities. The resulting error distributions are presented in Fig.~\ref{fig:speed_error_mse}.

As illustrated, FastStair achieves the lowest MAE (approximately $0.5\,\text{m/s}$), validating the effectiveness of the full training pipeline. Notably, the Pre-trained policy exhibits a slightly higher MAE than the Baseline. This result is consistent with our design intent: the heavy weighting of the foothold-tracking reward during pre-training biases the policy toward precise, safe foot placement (as evidenced in Fig.~\ref{fig:footstep_error}) at the expense of velocity-tracking performance. This safety-versus-agility trade-off explicitly justifies the necessity of the subsequent post-training stage to recover velocity-tracking precision while retaining the learned foothold priors.

Conversely, the AMP method yields the largest MAE. As noted in Sec.~\ref{sec:TraversalSuccessRatio}, the AMP policy frequently exhibits stagnation behavior at the stair base under high-velocity commands, leading to significant tracking deviations. Finally, we note that the maximum error values across all methods are inflated by velocity measurements recorded immediately prior to falls; therefore, maximum error is not a reliable indicator of tracking quality.
\begin{figure*}[htb]
    \centering
    \includegraphics[width=0.8\textwidth]{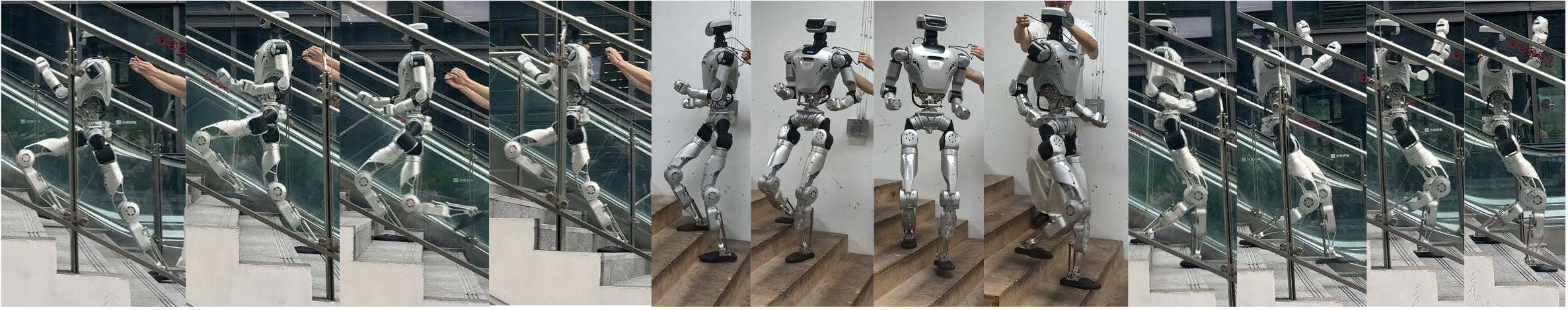}
    \setlength{\belowcaptionskip}{-10pt}  
    \caption{Outdoor experiments of perception-enabled agile stair climbing, including high-speed ascent and descent and turning maneuvers on stairs.}
    \label{fig:shiyanjietu}
\end{figure*}
\subsection{Real-World Deployment}



We deploy the final policy on the full-scale humanoid robot ``LimX Oli,'' executing directly on the onboard RK3588-based computer at a control frequency of $100\,\text{Hz}$, without any additional task-level fine-tuning. We conduct a series of outdoor experiments to validate performance (see Fig.~\ref{fig:robot_real_1} and Fig.~\ref{fig:shiyanjietu}). Detailed video recordings are provided in the supplementary material \footnote{\url{https://youtu.be/SoLBK7VEGDo}}.

We perform comprehensive stair-climbing tests covering the full velocity range (0 to 1.5 m/s) on an outdoor staircase with a 15 cm step height (Fig.~\ref{fig:shiyanjietu}). The robot successfully executes both ascent and descent across the full commanded velocity range, achieving ascent velocities up to 1.5 m/s. To accommodate such high-speed motion, the robot exhibits an emergent step-skipping behavior, clearing two steps in a single stride to maximize forward progress. Notably, the robot successfully negotiated 33 stairs (17 cm/step) at the Canton Tower within 12 seconds, reaching a peak velocity of 1.65 m/s during the final steps of the ascent.

Further validation includes the continuous negotiation of steeper staircases ($20\,\text{cm}$ step height, Fig.~\ref{fig:robot_real_1}) and dynamic turning maneuvers between flight segments. All tasks were completed successfully, demonstrating that FastStair enables stable, perception-aware agile locomotion in challenging real-world environments. 

\section{Conclusion}

In this paper, we presented FastStair, a multi-stage learning framework that reconciles model-based stability with learned agility for high-speed stair climbing. By reformulating DCM-based foothold optimization into a GPU-parallel discrete search, we provided real-time guidance within the RL loop with negligible computational overhead. To mitigate planner-induced conservatism and improve velocity-tracking performance, the  base policy was fine-tuned into separate low- and high-speed experts to accommodate differing action distributions. To eliminate instabilities associated with hard switching, the experts were merged into a single network and consolidated via LoRA fine-tuning, yielding a unified policy that operates robustly across the full commanded-speed range. Experimental results on the Oli robot demonstrate robust performance, achieving agile stair ascent at speeds up to 1.65 m/s. The successful navigation of a 33-step spiral staircase in 12 seconds underscores the efficiency and high-speed capabilities of FastStair in challenging real-world environments.

For future work, we aim to extend the parallel planning paradigm to a broader spectrum of challenging terrains. Furthermore, we plan to investigate generalized methodologies for tightly coupling model-based geometric priors with large-scale RL, ultimately enabling reliable deployment in complex, unstructured field environments.

\addtolength{\textheight}{-12cm}   









\bibliographystyle{IEEEtran} 
\bibliography{IEEEabrv,IEEEexample}

\end{document}